\pdfoutput=1

\documentclass[11pt]{article}

\usepackage[final]{acl}

\usepackage{times}
\usepackage{latexsym}

\usepackage[T1]{fontenc}

\usepackage[utf8]{inputenc}

\usepackage{microtype}

\usepackage{inconsolata}

\usepackage{graphicx}
\usepackage{xcolor}
\usepackage{url}
\usepackage{bm}
\usepackage{multirow}
\usepackage{amsmath,amssymb}
\usepackage{amsthm}
\usepackage{amsfonts}
\usepackage{tabularx}
\usepackage{paralist}
\usepackage{booktabs}
\usepackage{bbm}
\usepackage[most]{tcolorbox}

\newenvironment{nlgmethod}[1]
  {\par\smallskip\noindent\textbf{#1}\space}
  {\par\smallskip}
\tcbset{
  colframe=green!60!black,
  colback=green!5,
  coltitle=white,
  fonttitle=\bfseries,
  boxrule=0.8mm,
  arc=0.5mm,
  top=1mm, bottom=1mm, left=2mm, right=2mm,
}

\title{LCES: Zero-shot Automated Essay Scoring via Pairwise Comparisons Using Large Language Models}

\author{Takumi Shibata \and Yuichi Miyamura \\
  Deloitte Analytics R\&D, Deloitte Touche Tohmatsu LLC \\
  \texttt{\{takumi.shibata, yuichi.miyamura\}@tohmatsu.co.jp}}

\begin{document}
\maketitle
\begin{abstract}
   Recent advances in large language models (LLMs) have enabled zero-shot automated essay scoring (AES), providing a promising way to reduce the cost and effort of essay scoring in comparison with manual grading. However, most existing zero-shot approaches rely on LLMs to directly generate absolute scores, which often diverge from human evaluations owing to model biases and inconsistent scoring. To address these limitations, we propose LLM-based Comparative Essay Scoring (LCES), a method that formulates AES as a pairwise comparison task. Specifically, we instruct LLMs to judge which of two essays is better, collect many such comparisons, and convert them into continuous scores. Considering that the number of possible comparisons grows quadratically with the number of essays, we improve scalability by employing RankNet to efficiently transform LLM preferences into scalar scores. Experiments using AES benchmark datasets show that LCES outperforms conventional zero-shot methods in accuracy while maintaining computational efficiency. Moreover, LCES is robust across different LLM backbones, highlighting its applicability to real-world zero-shot AES.
\end{abstract}

\section{Introduction}\label{sec::introduction}
Automated essay scoring (AES) aims to assess the quality of written essays using natural language processing and machine learning techniques. AES has garnered significant attention as a means to reduce the cost relative to human grading and to ensure fairness~\citep{Uto2021Review,Do2023ProTACT}.

Most conventional AES methods focus on \textit{prompt-specific} approaches\footnote{Here, we use \textit{prompt} to refer to the essay topic and \textit{LLM prompt} to refer to instructions given to language models.}, which train machine learning models or neural networks on scored essays tailored to each essay prompt~\citep{Alikaniotis2016Automatic,Dong2017Attention,Yang2020R2BERT,Xie2022NPCR,Shibata2022Analytic,Wang2025T-MES}. However, this approach requires collecting large amounts of scored essays for every prompt, resulting in substantial costs. To address this issue, recent studies have proposed \textit{cross-prompt} AES methods that leverage domain adaptation or domain generalization techniques~\citep{Ridley2021CTS,Chen2023PMAES,Do2023ProTACT,Chen2024PLAES,Li2025KAES}. In those techniques, models are trained on scored essays from source prompts and evaluated on a different, target prompt. Although these methods can maintain high score accuracy even when scored essays for the target prompt are scarce or unavailable, they still require a certain amount of scored essay data for training, leaving unsolved the fundamental challenge of satisfying data requirements.

\begin{figure}
   \centering
   \includegraphics[width=7.7cm]{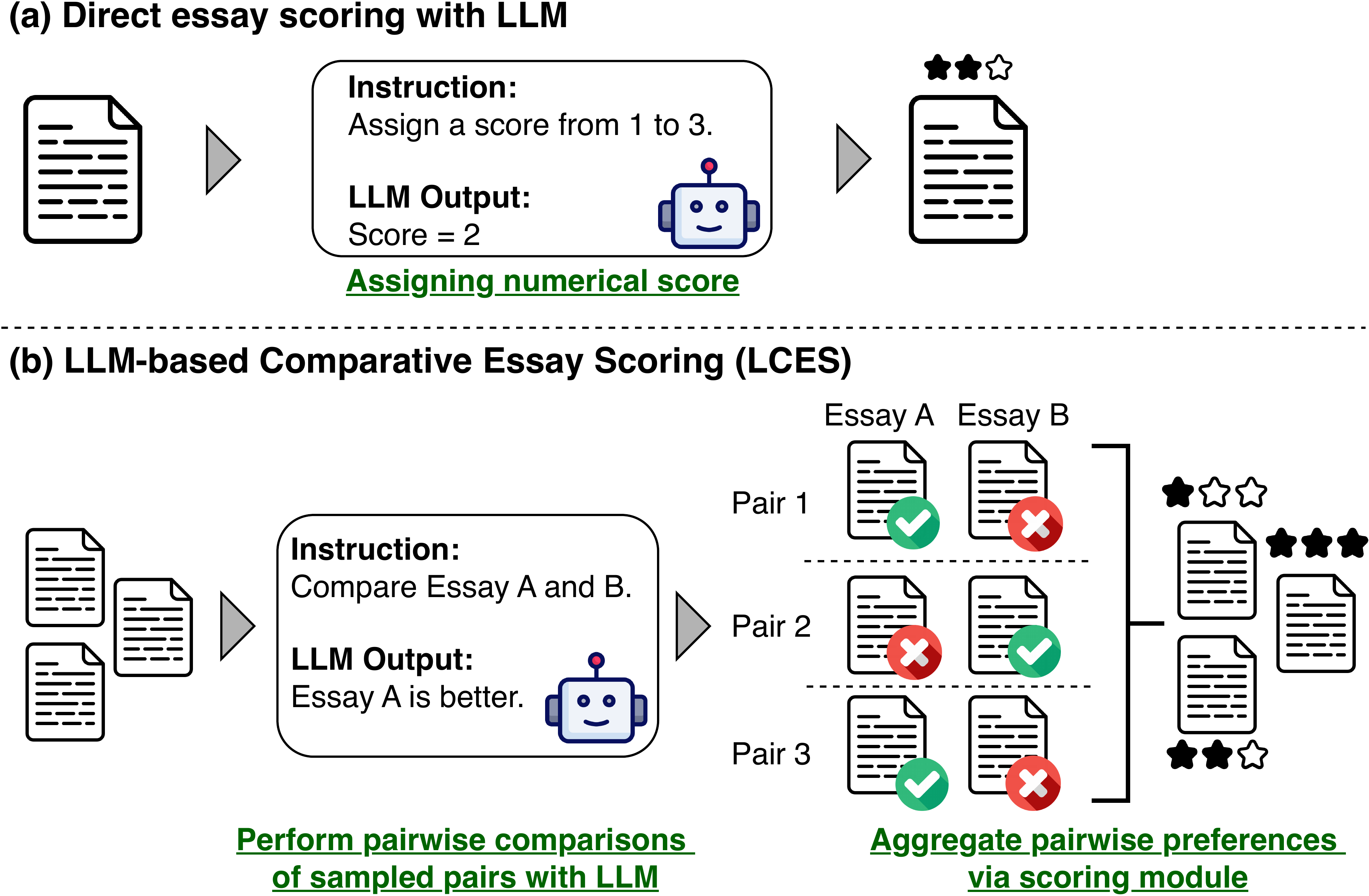}
   \caption{Comparison between (a) direct essay scoring using LLMs and (b) our proposed LCES framework.}
   \label{fig::essay}
\end{figure}

In parallel to the above, large language models (LLMs) have recently demonstrated remarkable capabilities across various natural language processing tasks in zero-shot settings~\citep{Kojima2022LLM}, motivating efforts to apply them to AES without the use of scored essays. A typical zero-shot AES approach instructs an LLM with a rubric and essay to generate a numerical score~\citep{Mizumoto2023Exploring}. A more advanced method first converts the original rubric defined in the dataset into trait-level rubrics using LLMs, then employs LLMs to independently predict scores for each trait, and finally aggregates these scores to estimate the overall score~\citep{lee-etal-2024-unleashing}. While these approaches are promising, they still have several limitations of direct score generation. They tend to be sensitive to the phrasing of LLM prompt and susceptible to model bias, and often exhibit grading behavior inconsistent with that of human raters~\citep{Zheng2023LLM-as-a-judge,Liu2024Pairs,Mansour2024Can,Li2025FromGeneration}.

Since such problems exist with direct scoring, this study explores an alternative evaluation method, namely pairwise evaluation. Instead of predicting absolute scores, we instruct the LLM to perform pairwise comparisons in which the better of two essays is determined. This approach is inspired by recent advances in LLM-based evaluation for natural language generation~\citep{Liu2024Pairs,Liusie2024LLM}, dialogue systems~\citep{Park2024PairEval}, and information retrieval~\citep{Qin2024LLM}, where pairwise comparisons have demonstrated stronger alignment with human preferences. Despite its promise, pairwise comparisons remain largely unexplored in the AES literature.

Against this backdrop, we propose \textbf{L}LM-based \textbf{C}omparative \textbf{E}ssay \textbf{S}coring (LCES), a novel framework for zero-shot AES that first collects pairwise comparisons using LLMs and then estimates continuous essay scores. As shown in Figure~\ref{fig::essay}, LCES differs from conventional LLM-based scoring by shifting from direct score generation to relative preference modeling. To scale this approach to large essay datasets, we employ RankNet~\citep{Burges2005RankNet}, which allows efficient training from pairwise comparisons without exhaustively enumerating all essay pairs. This mitigates the quadratic complexity in the number of items as is typically seen in pairwise comparisons~\citep{Liusie2024Efficient}.

Through comprehensive experiments using standard AES benchmark datasets, we demonstrate that LCES substantially outperforms existing zero-shot scoring methods. Moreover, LCES is robust to the choice of LLM and can be applied with virtually any model, making it well suited for practical deployment.

The contributions of this work are summarized as follows:
\begin{enumerate}[(1)]
   \item We introduce the first AES framework based on LLM-generated pairwise comparisons, addressing key limitations of direct score generation.
   \item We leverage RankNet to convert LLM-generated preferences into continuous scores, enabling accurate and computationally efficient zero-shot AES.
   \item Extensive experiments confirm that LCES outperforms conventional zero-shot AES baselines and is robust across different types of LLMs.
 \end{enumerate}

\section{Related Work}\label{sec::related_work}

\paragraph{Automated Essay Scoring.}
Early AES systems were largely \textit{prompt-specific}, beginning with handcrafted-feature-based models~\cite{Yannakoudakis2011New} and later adopting neural networks~\citep{Dong2017Attention,Xie2022NPCR}. Because it is costly to collect scored essays for every new prompt, \textit{cross-prompt} methods have been proposed to train models that generalize across prompts~\citep{Ridley2020PAES,Chen2023PMAES,Chen2024PLAES}. Recently, \textit{zero-shot} AES using LLMs has emerged~\cite{Mizumoto2023Exploring,Yancey2023rating,Wang2024Beyond,Mansour2024Can,lee-etal-2024-unleashing}, enabling score generation without the use of scored essays. \citet{Mizumoto2023Exploring} used \mbox{OpenAI}'s \texttt{text-davinci-003} to score essays based on rubric and essay content. In a zero-shot framework called Multi-Trait Specification (MTS), \citet{lee-etal-2024-unleashing} instructed an LLM to generate trait-level rubrics and then used them to evaluate essays by scoring each trait individually and aggregating the results. \citet{Mansour2024Can} demonstrated that LLM-generated scores are highly sensitive to the instructions given to the model, raising concerns about reliability. While zero-shot AES offers a promising direction, its scoring accuracy still lags behind supervised prompt-specific and cross-prompt methods.

\paragraph{LLM-based Evaluation.}
With the growing zero-shot capabilities of LLMs, the \textit{LLM-as-a-judge} paradigm~\citep{Zheng2023LLM-as-a-judge} has gained attention as a general framework for using LLMs in evaluation tasks. Although direct score generation is common, it often suffers from LLM prompt sensitivity~\citep{Li2025FromGeneration} and misalignment with human judgments~\citep{Liu2024Pairs}. To improve reliability, recent studies in natural language generation~\citep{Liu2024Pairs,Liusie2024LLM}, dialogue systems~\citep{Park2024PairEval}, and information retrieval~\citep{Qin2024LLM} have instructed LLMs to make pairwise comparisons in which the better of two candidates is selected. Compared with absolute scoring, this approach requires fewer reasoning steps by LLMs and yields more consistent and human-aligned judgments. However, it remains underexplored in AES.

\paragraph{Comparisons to Scores.}
Converting pairwise comparisons into continuous scores, which can be interpreted as latent measures of item quality that explain observed comparisons, has been widely studied. The Elo rating system~\cite{Elo1978} updates scores iteratively based on match outcomes. The Bradley--Terry model~\citep{BT1952} estimates win probabilities using the difference in latent scores between items, which are inferred by maximizing the likelihood of the observed comparisons. RankNet~\citep{Burges2005RankNet} extended this idea by learning latent scores from input features via a neural pairwise loss function. We use RankNet to transform LLM-generated essay comparisons into latent scores, enabling accurate and computationally efficient zero-shot AES.

\section{Proposed Method}

We start with a set of unscored essays $\mathcal{D} = \{ x_i \}_{i=1}^{N}$, where $x_i$ denotes the $i$th essay and $N$ is the total number of essays. The goal of LCES is to estimate a latent score $\hat{s}_i$ for each essay $x_i$, representing its relative quality within the set $\mathcal{D}$. Depending on the assessment objective, the estimated score $\hat{s}_i$ can be converted into a ranking $\hat{r}_i$ or a score $\hat{y}_i$ aligned with a predefined rubric.

LCES consists of three main steps: 
\begin{inparaenum}[(1)]
  \item \textbf{Pairwise comparison generation:} Sample essay pairs from $\mathcal{D}$ and use an LLM to judge which essay is better, or whether they are of equal quality, based on a given rubric; 
  \item \textbf{Latent score estimation:} Train a RankNet model on the comparison dataset to estimate a latent score $\hat{s}_i$ for each essay; and
  \item \textbf{Output conversion:} Convert the latent score $\hat{s}_i$ into either a ranking $\hat{r}_i$ or a score $\hat{y}_i$, depending on the evaluation goal.
\end{inparaenum}
Each step is described in detail in the following subsections.

\subsection{Pairwise Comparison Generation}\label{sec::generate_pairwise_comparison}
\begin{figure}[tb]
   \centering
   \includegraphics[width=7cm]{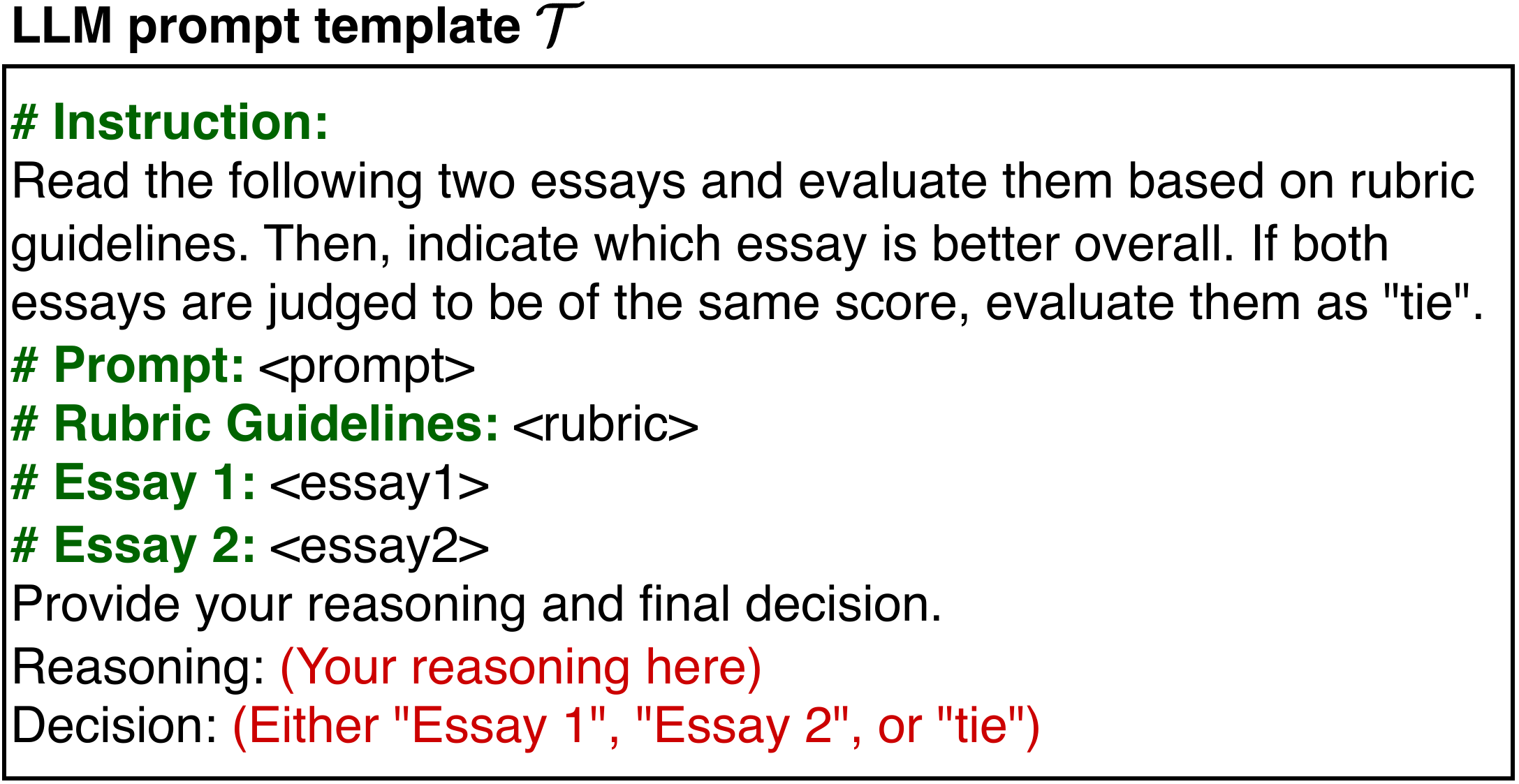}
   \caption{Simplified LLM prompt template $\mathcal{T}$ used for pairwise essay comparisons.}
   \label{fig::prompt}
\end{figure}
To generate pairwise comparisons, we use an LLM prompt template $\mathcal{T}$ that guides the LLM to evaluate two essays based on a given rubric. A simplified version of this template is shown in Figure~\ref{fig::prompt}, and the complete version can be found in Appendix~\ref{sec::llm_prompts}. Given essay prompt $p$, scoring rubric $r$, and two essays $x_i$ and $x_j$, we construct the query $\mathcal{T}(p, r, x_i, x_j)$ by inserting each input into the corresponding placeholder in the template. Specifically, the placeholders \texttt{<prompt>}, \texttt{<rubric>}, \texttt{<essay1>}, and \texttt{<essay2>} are replaced with $p$, $r$, $x_i$, and $x_j$ respectively. To improve the reliability and interpretability of the comparisons, we use chain-of-thought prompting~\cite{Wei2022CoT}. This encourages the LLM to explain its reasoning before making a final decision. For the essay judged to be better, the LLM outputs a categorical label $w_{ij}$, which is one of ``Essay 1'', ``Essay 2'', and ``tie''. We convert this to a numerical label $c_{ij}$ by assigning scores of 1, 0, and 0.5 for ``Essay 1'', ``Essay 2'', and ``tie'', respectively.

LLMs can be sensitive to the order in which the two essays are presented~\cite{Zheng2023LLM-as-a-judge}. To reduce this position bias, we query the LLM twice for each pair. One query presents the essays as $(x_i, x_j)$, and the other as $(x_j, x_i)$. Let $c_{ij}$ be the numerical label from the first query and $c_{ji}$ be the label from the second. We define the final debiased label $\tilde{c}_{ij}$ as follows:
\begin{equation}
\tilde{c}_{ij} = \begin{cases}
c_{ij} & \text{if}\  c_{ij} = 1 - c_{ji} \\
0.5 & \text{otherwise}.
\end{cases}
\end{equation}
If the two results are consistent, we retain the original label. If the results contradict each other or one of them indicates a tie, we treat the pair as a tie.

To apply this comparison procedure, we construct a set of essay pairs. Let $\mathcal{I} = \{(i, j) \mid i \neq j,\ i, j \in \{1, 2, \ldots, N\}\}$ be the set of all possible ordered essay pairs. Since comparing all $N(N - 1)$ pairs is computationally expensive, we randomly sample a subset $\mathcal{I}_s \subset \mathcal{I}$ containing $M$ pairs, where $M \ll N(N - 1)$. For each sampled pair, we obtain a debiased label $\tilde{c}_{ij}$ as described above. This yields the pairwise comparison dataset $\mathcal{D}_{\text{pair}} = \{(x_i, x_j, \tilde{c}_{ij}) \mid (i, j) \in \mathcal{I}_s\}$, which is used to train the RankNet model.

\subsection{Latent Score Estimation}\label{sec::ranknet}

\begin{figure}[tb]
   \centering
   \includegraphics[width=7cm]{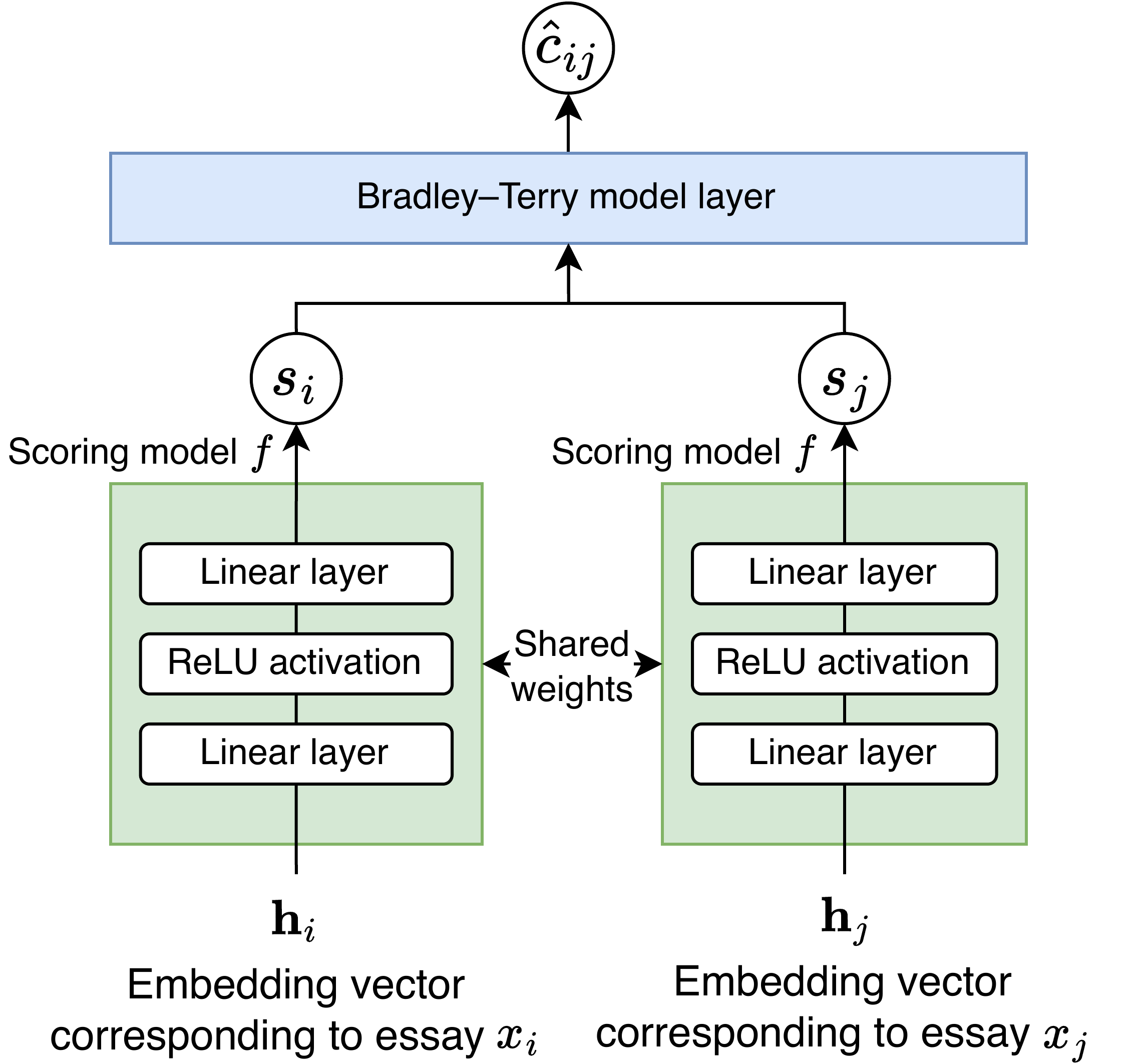}
   \caption{Architecture of the RankNet model used to estimate latent essay scores $\hat{s}_i$ from pairwise comparisons.}
   \label{fig::model}
\end{figure}

Using the pairwise comparison dataset $\mathcal{D}_{\text{pair}}$ generated in Section~\ref{sec::generate_pairwise_comparison}, we estimate a latent score $\hat{s}_i$ for each essay $x_i$. To this end, we employ RankNet~\citep{Burges2005RankNet}, a neural model designed to learn latent scores from pairwise preferences.

As shown in Figure~\ref{fig::model}, RankNet uses two parallel multi-layer perceptrons (MLPs) with shared weights. These form a scoring model, denoted by $f$, which maps an input essay representation to a scalar score. Specifically, we first convert each essay $x_i$ into an embedding vector $\bm{h}_i$ using any suitable text embedding model, and then compute its score as $s_i = f(\bm{h}_i)$. Each MLP consists of two linear layers with a ReLU activation~\citep{Agarap2019ReLU} applied after the first layer.

Given two essays $x_i$ and $x_j$, the model computes scores $s_i$ and $s_j$ using the shared network $f$, and estimates the probability that $x_i$ is preferred over $x_j$ as:
\begin{equation}
\hat{c}_{ij} = \sigma(s_i - s_j) = \frac{1}{1 + \exp(-(s_i - s_j))}.
\end{equation}
Here, $\sigma(\cdot)$ denotes the sigmoid function. This formulation mirrors the Bradley--Terry model~\citep{BT1952} and enables probabilistic modeling of pairwise preferences.

The model is trained to minimize the discrepancy between the predicted preference $\hat{c}_{ij}$ and the debiased target label $\tilde{c}_{ij}$. We use the binary cross-entropy loss $\mathcal{L}$:

{\small
\begin{equation} \label{eq:bce_loss}
   \mathcal{L} = -\frac{1}{M} \sum_{(i, j) \in \mathcal{I}_s} \left[ \tilde{c}_{ij} \log \hat{c}_{ij} + (1 - \tilde{c}_{ij}) \log (1 - \hat{c}_{ij}) \right]. \nonumber
\end{equation}
}
Let $\mathcal{S} = \{s_i\}_{i=1}^{N}$ denote the set of latent essay scores. The optimized scores $\hat{\mathcal{S}} = \{\hat{s}_i\}_{i=1}^{N}$ are obtained by minimizing the loss: $\hat{\mathcal{S}} = \arg\min_{\mathcal{S}} \mathcal{L}$.

\subsection{Output Conversion}\label{sec:output}

The estimated latent scores $\hat{s}_i$ can be converted into standard AES outputs, such as numerical scores or rankings, depending on the evaluation goal.

To produce a score $\hat{y}_i$ within a rubric-defined range $[y_{\min}, y_{\max}]$, we apply a linear transformation to the latent scores:

\begin{equation}
   \hat{y}_i = \frac{\hat{s}_i - s_{\min}}{s_{\max} - s_{\min}} \times (y_{\max} - y_{\min}) + y_{\min},
\end{equation}
where $s_{\min} = \min_i \hat{s}_i$ and $s_{\max} = \max_i \hat{s}_i$ are the minimum and maximum latent scores across all essays. If the rubric defines discrete score levels, the resulting $\hat{y}_i$ can optionally be rounded to the nearest valid level.

Alternatively, a ranking $\hat{r}_i$ can be obtained by sorting essays in descending order of their latent scores $\hat{s}_i$. This is useful in settings where only relative essay quality is required.

\section{Experiments}\label{sec::experiments}
We empirically evaluate the effectiveness of LCES through experiments using AES benchmark datasets, focusing on scoring performance and comparisons with existing methods.

\subsection{Datasets}

We utilized the following two benchmark datasets, which are commonly used in AES research~\cite{Taghipour2016Neural,Chen2023PMAES,lee-etal-2024-unleashing,Wang2024Beyond}:
\begin{nlgmethod}{ASAP (Automated Student Assessment Prize)}
   is a dataset released by the Kaggle competition\footnote{https://www.kaggle.com/c/asap-aes}. It consists of 12,978 essays across eight different prompts, each with human-assigned scores.
\end{nlgmethod}
\begin{nlgmethod}{TOEFL11}
   is a dataset of essays written by non-native English speakers taking the TOEFL iBT~\citep{Blanchard2013TOEFL11}. It contains 12,100 essays across eight different prompts, each with human-assigned scores.
\end{nlgmethod}

Table~\ref{tab:asap} summarizes the statistics of the ASAP and TOEFL11 datasets.
\begin{table}[tb]
   \caption{Statistics of the ASAP and TOEFL11 datasets. l/m/h denotes low/medium/high.}
   \label{tab:asap}
   \centering
   \small
   \resizebox{0.45\textwidth}{!}{
   \begin{tabular}{lcccc}
   \toprule
   \textbf{Dataset} & \textbf{Prompt} & \textbf{No. of Essays} & \textbf{Avg. Len.} & \textbf{Score Range}\\
   \midrule
   \multirow{8}{*}{ASAP} & 1 & 1,783 & 427 & 2--12\\ 
    & 2 & 1,800 & 432 & 1--6\\
    & 3 & 1,726 & 124 & 0--3\\
    & 4 & 1,772 & 106 & 0--3\\
    & 5 & 1,805 & 142 & 0--4\\
    & 6 & 1,800 & 173 & 0--4\\
    & 7 & 1,569 & 206 & 0--30\\
    & 8 & 723 & 725 & 0--60\\
   \midrule
   \multirow{8}{*}{TOEFL11} & 1 & 1,656 & 342 & l/m/h\\
    & 2 & 1,562 & 361 & l/m/h\\
    & 3 & 1,396 & 346 & l/m/h\\
    & 4 & 1,509 & 340 & l/m/h\\
    & 5 & 1,648 & 361 & l/m/h\\
    & 6 & 960 & 360 & l/m/h\\
    & 7 & 1,686 & 339 & l/m/h\\
    & 8 & 1,683 & 344 & l/m/h\\
   \bottomrule
   \end{tabular}%
   }
 \end{table}

\subsection{Baselines}
We adopted the following two zero-shot AES methods as baselines for comparison with LCES:
\begin{nlgmethod}{Vanilla.}
   A direct scoring approach where the LLM generates a rubric-aligned score for each essay without pairwise comparison. It uses chain-of-thought prompting to elicit reasoning before scoring. We used the same LLM prompts and hyperparameters as \citet{lee-etal-2024-unleashing}.
\end{nlgmethod}
\begin{nlgmethod}{MTS.}
   As described in Section~\ref{sec::related_work}, MTS~\citep{lee-etal-2024-unleashing} is a state-of-the-art zero-shot AES framework. The original implementation used GPT-3.5 to generate trait-level rubrics from the original rubric. In our experiments, we used GPT-4o instead because GPT-3.5 is no longer available. All other LLM prompts and hyperparameters followed the original implementation.
\end{nlgmethod}
\begin{table*}[tb]
   \centering
   \tiny
   \caption{QWK scores for each essay prompt in ASAP and TOEFL11. \textbf{Bold} indicates the best-performing method for each prompt. \textbf{P1-8} refers to Prompt 1 through Prompt 8.}
   \label{tab:asap_toefl}
   \resizebox{0.94\textwidth}{!}{
   \begin{tabular}{l l l c c c c c c c c l}
       \toprule
       \textbf{Dataset} & \textbf{Model} & \textbf{Method} & \textbf{P1} & \textbf{P2} & \textbf{P3} & \textbf{P4} & \textbf{P5} & \textbf{P6} & \textbf{P7} & \textbf{P8} & \textbf{Avg.}\\
       \midrule
       \multirow{15}{*}{ASAP} & Mistral-7B & Vanilla & 0.429     & 0.439     & 0.387     & 0.518     & 0.576     & 0.534     & 0.276     & 0.209     & 0.429 \\
       & & MTS     & 0.546     & 0.479     & 0.481     & \textbf{0.683}     & 0.706     & 0.519     & 0.501     & 0.175     & 0.511 \\
       & & LCES    & \textbf{0.600}     & \textbf{0.603}     & \textbf{0.690}     & 0.614     & \textbf{0.729}     & \textbf{0.792}     & \textbf{0.591}     & \textbf{0.315}     & \textbf{0.617} \\
       \cmidrule{2-12}
       & Llama-3.2-3B & Vanilla & 0.254 & 0.405 & 0.410 & 0.009 & 0.397 & 0.330 & 0.438 & 0.276 & 0.315 \\
       & & MTS     & 0.197 & 0.452 & 0.353 & 0.507 & 0.460 & 0.462 & 0.146 & 0.190 & 0.346 \\
       & & LCES    & \textbf{0.555} & \textbf{0.608} & \textbf{0.647} & \textbf{0.603} & \textbf{0.717} & \textbf{0.756} & \textbf{0.580} & \textbf{0.612} & \textbf{0.635} \\
       \cmidrule{2-12}
       & Llama-3.1-8B & Vanilla & 0.129     & 0.023     & 0.243     & 0.550     & 0.301     & 0.341     & 0.006     & -0.042     & 0.194 \\
       & & MTS     & 0.516     & 0.483     & 0.284     & 0.461     & 0.479     & 0.378     & 0.328     & 0.199     & 0.391 \\
       & & LCES    & \textbf{0.669} & \textbf{0.599} & \textbf{0.662} & \textbf{0.651} & \textbf{0.710} & \textbf{0.707} & \textbf{0.727} & \textbf{0.636} & \textbf{0.670} \\
       \cmidrule{2-12}
       & GPT-4o-mini & Vanilla & 0.106 & 0.402 & 0.314 & 0.602 & 0.577 & 0.470 & 0.425 & 0.517 & 0.426 \\
       & & MTS     & 0.472 & 0.386 & 0.448 & 0.552 & 0.708 & 0.419 & 0.479 & 0.412 & 0.485 \\
       & & LCES    & \textbf{0.537} & \textbf{0.602} & \textbf{0.679} & \textbf{0.638} & \textbf{0.709} & \textbf{0.737} & \textbf{0.614} & \textbf{0.521} & \textbf{0.630} \\
       \cmidrule{2-12}
       & GPT-4o & Vanilla & 0.216 & 0.498 & 0.447 & 0.681 & 0.710 & 0.571 & 0.535 & 0.411 & 0.509 \\
       & & MTS     & 0.380 & 0.547 & 0.513 & 0.621 & 0.500 & 0.515 & 0.421 & 0.432 & 0.491 \\
       & & LCES    & \textbf{0.531} & \textbf{0.592} & \textbf{0.702} & \textbf{0.626} & \textbf{0.747} & \textbf{0.766} & \textbf{0.669} & \textbf{0.593} & \textbf{0.653} \\
       \midrule
       \multirow{15}{*}{TOEFL11} & Mistral-7B & Vanilla & 0.235 & 0.128 & 0.174 & 0.106 & 0.050 & 0.046 & 0.106 & 0.222 & 0.133 \\
       & & MTS     & \textbf{0.634} & 0.496 & 0.571 & \textbf{0.607} & \textbf{0.603} & \textbf{0.573} & \textbf{0.578} & \textbf{0.689} & \textbf{0.594} \\
       & & LCES    & 0.415 & \textbf{0.514} & \textbf{0.663} & 0.519 & 0.508 & 0.496 & 0.532 & 0.644 & 0.536 \\
       \cmidrule{2-12}
       & Llama-3.2-3B & Vanilla & 0.184 & 0.117 & 0.291 & 0.195 & 0.149 & 0.206 & 0.067 & 0.149 & 0.170 \\
       & & MTS     & 0.361 & 0.389 & 0.454 & 0.456 & 0.341 & 0.364 & 0.323 & 0.299 & 0.373 \\
       & & LCES    & \textbf{0.615} & \textbf{0.542} & \textbf{0.709} & \textbf{0.678} & \textbf{0.582} & \textbf{0.479} & \textbf{0.555} & \textbf{0.708} & \textbf{0.608} \\
       \cmidrule{2-12}
       & Llama-3.1-8B & Vanilla & -0.036 & 0.148 & 0.003 & 0.021 & 0.019 & -0.023 & -0.029 & 0.063 & 0.021 \\
       & & MTS     & 0.368 & 0.408 & 0.407 & 0.311 & 0.351 & 0.285 & 0.335 & 0.379 & 0.356 \\
       & & LCES    & \textbf{0.597} & \textbf{0.570} & \textbf{0.727} & \textbf{0.697} & \textbf{0.652} & \textbf{0.550} & \textbf{0.558} & \textbf{0.717} & \textbf{0.633} \\
       \cmidrule{2-12}
       & GPT-4o-mini & Vanilla & 0.094     & 0.202     & 0.182     & 0.107     & 0.041     & 0.101     & 0.126     & 0.124     & 0.122 \\
       & & MTS     & 0.439 & 0.529 & 0.548 & 0.521 & 0.603 & 0.501 & 0.536 & 0.591 & 0.533 \\
       & & LCES    & \textbf{0.655} & \textbf{0.559} & \textbf{0.722} & \textbf{0.692} & \textbf{0.633} & \textbf{0.649} & \textbf{0.629} & \textbf{0.724} & \textbf{0.658} \\
       \cmidrule{2-12}
       & GPT-4o & Vanilla & 0.206     & 0.208     & 0.365     & 0.189     & 0.211     & 0.245     & 0.226     & 0.252     & 0.238 \\
       & & MTS     & 0.480 & 0.539 & 0.607 & 0.545 & 0.469 & 0.526 & 0.426 & 0.664 & 0.532 \\
       & & LCES    & \textbf{0.604} & \textbf{0.545} & \textbf{0.734} & \textbf{0.671} & \textbf{0.713} & \textbf{0.572} & \textbf{0.580} & \textbf{0.739} & \textbf{0.645} \\
       \bottomrule
   \end{tabular}
   }
 \end{table*}
\subsection{Experimental Setup}\label{sec::experimental_setup}

\begin{nlgmethod}{LLMs.}
   We conducted our evaluation using five distinct LLMs, namely, Mistral-7B (-instruct-v0.2)~\citep{jiang2023mistral7b}, Llama-3.2-3B (-Instruct), Llama-3.1-8B (-Instruct)~\citep{grattafiori2024llama3herdmodels}, GPT-4o-mini (-2024-07-18), and GPT-4o (-2024-08-06)~\citep{openai2024gpt4ocard}. All LLM inferences were performed with a temperature setting of 0.1.
\end{nlgmethod}

\begin{nlgmethod}{Implementation Details.}
   The number of sampled pairwise comparisons $M$ was set to 5{,}000 to construct the $\mathcal{D}_{\text{pair}}$ dataset. Essay embedding vectors $\bm{h}_i$ were generated using OpenAI's \texttt{text-embedding-3-large} model. Results obtained with alternative embedding models are presented in Appendix~\ref{sec::embedding_model}. The RankNet model was trained for 100 epochs using the Adam~\cite{Kingma2014Adam} optimizer with a learning rate of 0.001. The full set of hyperparameters is provided in Appendix~\ref{sec::hyperparameters}.
\end{nlgmethod}

\begin{nlgmethod}{Rubrics.}
   For pairwise comparisons within the ASAP dataset, we used the original scoring rubrics provided with the dataset. For the TOEFL11 dataset, consistent with previous studies~\citep{Mizumoto2023Exploring,lee-etal-2024-unleashing}, we used the IELTS Task 2 Writing Band Descriptors as the evaluation rubric.
\end{nlgmethod}

\begin{nlgmethod}{Evaluation Metrics.}
   We evaluated model performance using two standard metrics in AES, namely quadratic weighted kappa (QWK)~\citep{Cohen1960QWK} and the Spearman rank correlation coefficient. Following common practice in previous work~\citep{Taghipour2016Neural,Alikaniotis2016Automatic,Dong2017Attention,Do2023ProTACT}, we primarily report QWK. Results for Spearman correlations are provided in Appendix~\ref{sec::spearman_correlation}. For the ASAP dataset, we followed \citet{lee-etal-2024-unleashing} and randomly sampled 10\% of essays from each prompt for evaluation. For TOEFL11, we used the predefined test split consisting of 1,100 essays across eight prompts.
\end{nlgmethod}

\begin{nlgmethod}{Scoring Strategy.}
   For QWK-based evaluation, we rounded the predicted scores $\hat{y}_i$ to align with the score range of each prompt in the ASAP dataset. For the TOEFL11 dataset, we first converted the latent scores to a $[1, 5]$ scale by the linear transformation described in Section~\ref{sec:output}, and we then mapped them to low/medium/high categories using thresholds of 2.25 and 3.75, following the approach used in previous research~\citep{Blanchard2013TOEFL11,lee-etal-2024-unleashing}.
\end{nlgmethod}

\subsection{Results and Discussion}\label{sec:results_discussion}

The results in Table~\ref{tab:asap_toefl} show that LCES outperforms both MTS and Vanilla in most settings, achieving higher average QWK scores across models and prompts, with particularly large gains on the ASAP dataset. The only exception is TOEFL11 with Mistral-7B, where LCES performs worse than MTS. As shown later in Section~\ref{sec:position_bias}, Mistral-7B exhibits a high inconsistency rate (51.4\%) when essay order is reversed, suggesting that it has difficulty in reliably identifying the better essay. Notably, Mistral-7B achieves the highest performance under the MTS setting on TOEFL11, surpassing more recent or larger models such as GPT-4o and Llama-3.1-8B. This suggests that MTS and LCES may favor different model capabilities. While LCES underperforms MTS with Mistral-7B, it consistently outperforms MTS with all other LLMs, highlighting the general effectiveness of the LCES framework.

Moreover, LCES exhibited lower performance variance across different LLM backbones compared to conventional methods. On ASAP, the standard deviation of its average performance across five backbone models is just 0.021, compared to 0.072 for MTS and 0.122 for Vanilla. On TOEFL11, LCES similarly shows low variability, with a standard deviation of 0.048 across models, outperforming MTS (0.106) and Vanilla (0.079). These low inter-model variances indicate that LCES remains stable regardless of backbone choice, whereas MTS and Vanilla fluctuate more, making their performance less predictable.

\begin{table}[tb]
   \centering
   \small
   \caption{Average QWK scores across all ASAP prompts for LCES and supervised learning baselines.}
   \label{tab:performance_comparison_qwk_suppervised}
   \resizebox{0.48\textwidth}{!}{
   \begin{tabular}{lc}
   \toprule
   \textbf{Method} & \textbf{Avg. QWK} \\
   \midrule
   \multicolumn{2}{l}{\textit{Prompt-specific}} \\
   NPCR~\citep{Xie2022NPCR} & 0.792 \\
   BERT-base-uncased~\citep{Devlin2019BERT} & 0.740 \\
   RoBERTa-base~\citep{Liu2019RoBERTaAR} & 0.743 \\
    \midrule
   \multicolumn{2}{l}{\textit{Cross-prompt}} \\
   PAES~\citep{Ridley2020PAES} & 0.678 \\
   PMAES~\citep{Chen2023PMAES} & 0.658 \\
   \midrule
   \multicolumn{2}{l}{\textit{Zero-shot}} \\
   LCES (Llama-3.1-8B) & 0.670 \\
   \bottomrule
   \end{tabular}
   }
 \end{table}
\section{Analysis}
We present a set of analyses to further examine the effectiveness and properties of the proposed framework beyond overall performance metrics.

\subsection{Comparison with Supervised Models}
Although LCES is a zero-shot method, we also compare it with several supervised learning baselines on the ASAP dataset, as summarized in Table~\ref{tab:performance_comparison_qwk_suppervised}. We include both prompt-specific and cross-prompt models. The prompt-specific models are trained on 90\% of the essays from a single prompt and evaluated on the remaining 10\%, using the same evaluation split described in Section~\ref{sec::experimental_setup}. The cross-prompt models are trained on essays from all prompts except the one under evaluation, and are also evaluated on the same 10\% split of the target prompt.
 
Specifically, we make comparisons against NPCR~\citep{Xie2022NPCR}, which is reported to provide state-of-the-art results on ASAP, as well as BERT~\citep{Devlin2019BERT} and RoBERTa~\citep{Liu2019RoBERTaAR} fine-tuned on the same prompt-specific splits. We also include PAES~\citep{Ridley2020PAES} and PMAES~\citep{Chen2023PMAES}, which are two strong cross-prompt baselines.

As shown in Table~\ref{tab:performance_comparison_qwk_suppervised}, LCES with Llama-3.1-8B, which achieved the highest overall performance among all tested LLMs in the zero-shot experiments (see Section~\ref{sec:results_discussion}), obtains QWK scores that are comparable to several supervised learning models. While NPCR, BERT, and RoBERTa still outperform LCES, the performance gap has significantly narrowed in comparison with previously reported zero-shot methods. In addition, LCES with Llama-3.1-8B achieves performance on par with the strong cross-prompt baselines PAES and PMAES\footnote{PMAES was run with a smaller batch size due to GPU limitations (RTX 4090), which may have led to reduced performance.}. Indeed, a Wilcoxon signed-rank test revealed no statistically significant differences in QWK scores between LCES and either PAES or PMAES ($p$-values all above the 0.05 significance threshold). This level of performance is unprecedented among zero-shot AES methods. These results highlight the effectiveness of the proposed method in the absence of scored essays.

\begin{table}[tb]
   \centering
   \small
   \caption{Average percentage of LLM judgments that change when the order of essay pairs is reversed, computed across all prompts in each dataset.}
   \label{tab:inconsistent_rates}
   \begin{tabular}{lcc}
   \toprule
   \textbf{Model} & \textbf{ASAP (\%)} & \textbf{TOEFL11 (\%)} \\
   \midrule
   Mistral-7B       & 42.8 & 51.4    \\
   Llama-3.2-3B     & 28.8 & 39.0 \\
   Llama-3.1-8B     & 21.6 & 23.8 \\
   GPT-4o-mini      & 13.8 & 10.5 \\
   GPT-4o           & 10.4 & 17.0 \\
   \bottomrule
   \end{tabular}
\end{table}
\subsection{Position Bias}\label{sec:position_bias}
We measure the impact of position bias by calculating the percentage of pairwise comparisons that change when the order of essays is reversed. Table~\ref{tab:inconsistent_rates} shows the inconsistency rates for each LLM on the same comparison pairs used to construct $\mathcal{D}_{\text{pair}}$ for ASAP and TOEFL11. As expected, larger models such as GPT-4o exhibit lower inconsistency, suggesting greater robustness to position bias. In contrast, Mistral-7B shows a particularly high inconsistency rate of 51.4\% on TOEFL11, indicating substantial sensitivity to essay order.

\begin{table}[t]
   \centering
   \small
   \caption{Average QWK on ASAP and TOEFL11 with and without position bias correction.}
   \label{tab:ablation_position}
   \resizebox{0.45\textwidth}{!}{
   \begin{tabular}{llccc}
   \toprule
   \multirow{2}{*}{\textbf{Dataset}} & \multirow{2}{*}{\textbf{Model}} & \multicolumn{2}{c}{\textbf{Avg. QWK}} \\ 
   \cmidrule{3-4}
   & & \textbf{w/o Debias} & \textbf{w/ Debias} \\
   \midrule
   \multirow{5}{*}{ASAP} & Mistral-7B    & 0.611 & 0.617 \\
                          & Llama-3.2-3B & 0.630 & 0.635 \\
                          & Llama-3.1-8B & 0.661 & 0.670 \\
                          & GPT-4o-mini  & 0.633 & 0.630 \\
                          & GPT-4o       & 0.649 & 0.653 \\
   \midrule
   \multirow{5}{*}{TOEFL11} & Mistral-7B   & 0.510 & 0.536 \\
                            & Llama-3.2-3B & 0.588 & 0.608 \\
                            & Llama-3.1-8B & 0.628 & 0.633 \\
                            & GPT-4o-mini  & 0.664 & 0.658 \\
                            & GPT-4o       & 0.648 & 0.645 \\
   \bottomrule
   \end{tabular}
   }
\end{table}
\begin{figure*}[t]
   \centering
   \includegraphics[width=0.94\textwidth]{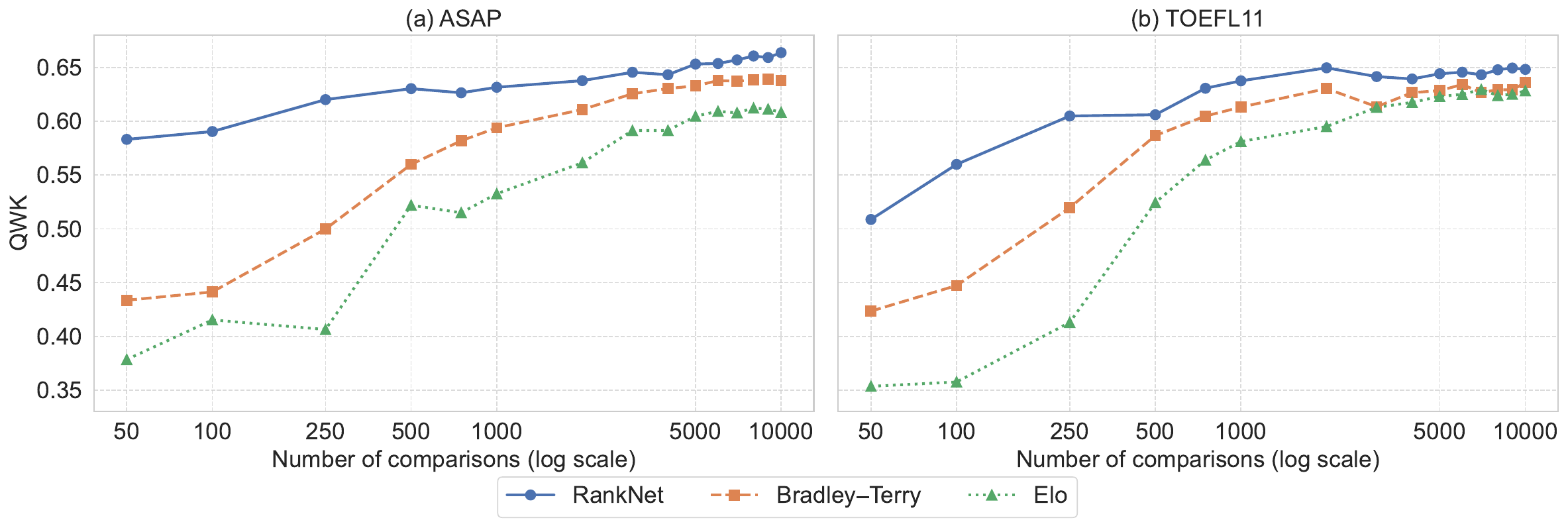}
   \caption{Relationship between the number of pairwise comparisons (log scale) and QWK scores. (a) ASAP dataset. (b) TOEFL11 dataset.}
   \label{fig:number_of_comparisons}
\end{figure*}
\begin{table*}[tb]
   \centering
   \small
   \caption{QWK scores of LCES in \textit{transductive} and \textit{inductive} settings, evaluated using GPT-4o-mini.}
   \label{tab:inductive_performance}
   \resizebox{0.84\textwidth}{!}{
   \begin{tabular}{llccccccccc}
       \toprule
       \textbf{Dataset} & \textbf{Setting} & \textbf{P1} & \textbf{P2} & \textbf{P3} & \textbf{P4} & \textbf{P5} & \textbf{P6} & \textbf{P7} & \textbf{P8} & \textbf{Avg.}\\
       \midrule
       \multirow{2}{*}{ASAP} & Transductive & 0.537 & 0.602 & 0.679 & 0.638 & 0.709 & 0.737 & 0.614 & 0.521 & 0.630 \\
        & Inductive    & 0.611 & 0.622 & 0.588 & 0.631 & 0.707 & 0.783 & 0.487 & 0.603 & 0.629 \\
       \midrule
       \multirow{2}{*}{TOEFL11} & Transductive & 0.655 & 0.559 & 0.722 & 0.692 & 0.633 & 0.649 & 0.629 & 0.724 & 0.658 \\
        & Inductive    & 0.624 & 0.613 & 0.715 & 0.617 & 0.622 & 0.615 & 0.616 & 0.707 & 0.641 \\
       \bottomrule
   \end{tabular}
   }
 \end{table*}
 To assess the effect of position bias correction, we compare average QWK scores with and without the position bias correction. As shown in Table~\ref{tab:ablation_position}, models with higher inconsistency rates, such as Mistral-7B and Llama-3.2-3B, tend to benefit more from the correction. These results suggest that the proposed correction method is generally more effective for models with higher position inconsistency, whereas its effect is limited for models that already exhibit low inconsistency.

\subsection{Comparison of Latent Score Conversion Methods}

We evaluate the effectiveness of different latent score conversion techniques by comparing our RankNet-based approach with the Bradley--Terry model and the Elo rating system which are representative methods described in Section~\ref{sec::related_work}. The experiment examines how the number of pairwise comparisons $M$, ranging from 50 to 10,000, affects scoring accuracy, measured by QWK, on the ASAP and TOEFL11 datasets. This experiment adopts GPT-4o as the LLM, in view of its robust performance on both datasets.

Figure~\ref{fig:number_of_comparisons} illustrates the performance trends. As $M$ increases, accuracy improves for all methods, highlighting the benefit of additional preference data. Among them, the RankNet-based approach consistently outperforms both the Bradley--Terry model and the Elo rating system across the entire range of $M$ on both datasets. Notably, RankNet achieves high QWK scores even with relatively few comparisons (e.g., $M=50$ or $M=100$), demonstrating strong performance particularly in limited data scenarios. This advantage likely stems from RankNet's ability to incorporate textual features directly from essays, whereas the baseline methods rely solely on comparison outcomes.

These results suggest that RankNet is highly effective for pairwise-based essay scoring. Its superior accuracy and greater data efficiency make it well suited for practical settings where collecting extensive comparison data may be costly or infeasible.

\subsection{Performance in the Inductive Setting}

In this section, we evaluate the performance of LCES under different problem settings. Throughout this paper, we have considered a problem setting where all target essays are available at once for scoring--we refer to this as the \textit{transductive} setting. However, another scenario is possible where new essays arrive individually for scoring after the model has been trained--we refer to this as the \textit{inductive} setting. The key advantage of LCES in the inductive setting is that the scoring model $f$, learned during RankNet training, maps essay embeddings to scalar scores and can thus generalize to unseen essays without requiring additional pairwise comparisons.

To simulate this scenario, we train the scoring model $f$ within the LCES framework on pairwise comparisons constructed from 90\% of the essays in each dataset (ASAP or TOEFL11), and then use it to predict scores for the remaining 10\%. We use GPT-4o-mini for its computational efficiency and low API cost.

QWK scores in the inductive setting are close to those in the transductive setting, with 0.629 vs.\ 0.630 on ASAP and 0.641 vs.\ 0.658 on TOEFL11 (Table~\ref{tab:inductive_performance}). These results demonstrate that $f$ generalizes effectively to unseen essays. This ability to score new essays without constructing additional comparisons involving them makes LCES well suited for inductive scenarios. In contrast, models such as the Bradley--Terry model or Elo require the generation of new comparisons for each essay, leading to higher deployment overhead in inductive settings.

\section{Conclusion}

In this study, we presented LLM-based Comparative Essay Scoring (LCES), a zero-shot AES framework that leverages LLM-driven pairwise comparisons to address key limitations of direct score generation. LCES instructs an LLM to judge which of two essays is better, and then trains a RankNet model to estimate continuous essay scores.

Experimental results on two benchmark datasets, namely, ASAP and TOEFL11, demonstrate that LCES consistently outperforms existing zero-shot methods in scoring accuracy. It maintains strong performance even with a limited number of comparisons and is robust to the choice of LLM. Moreover, LCES can be applied in inductive settings without requiring additional comparisons for new essays. These properties make LCES well suited for real-world AES applications.

\section*{Limitations}

Despite its advantages, LCES has several limitations. First, it relies on pairwise preference labels generated by an LLM, which may contain noise or inconsistencies. These imperfect labels directly affect the quality of learned scoring model $f$.

Second, while LCES tends to perform reliably when provided a sufficient number of comparisons $M$, it remains unclear how to determine an appropriate value of $M$. This limits the ability to systematically control scoring quality.

Third, LCES maps latent relative scores to an absolute scale via linear transformation, assuming sampled comparisons span the full score range. If low- or high-scoring essays are missing, the transformation may yield inaccurate absolute scores. While ranking performance would remain unaffected, this can reduce alignment with human judgment in tasks requiring precise or rubric-specific scoring.

Finally, the zero-shot nature of LCES means that, without labeled data, its performance cannot be quantitatively assessed. For practical deployment, caution is warranted when used in high-stakes exams.

\section*{Acknowledgments}
This research was conducted as part of an internal project at Deloitte Touche Tohmatsu LLC. The authors would like to thank Tomoaki Geka and Tomotake Kozu for helpful discussions and support. We also used ChatGPT to assist in clarifying the structure and wording of certain parts of the manuscript.

\section*{Ethics Statement}
The primary ethical consideration for LCES is its reliance on LLMs for pairwise essay comparisons. LLMs may inherit and perpetuate biases present in their extensive training data. While our method includes a debiasing step for position bias, other latent biases could potentially influence the fairness of evaluations across different student demographics or writing styles. We recommend further auditing for such biases before any high-stakes deployment of LCES.

The experiments in this study were conducted using publicly available and established benchmark datasets (ASAP and TOEFL11). No new personally identifiable information was collected or used in this research.

We envision LCES as an assistive tool to support human graders, reducing their workload and potentially improving consistency, rather than as a complete replacement for human judgment. Given its zero-shot nature, thorough validation of LCES on specific target prompts and scoring rubrics is crucial before its application in real-world educational assessments to ensure reliability and prevent potential negative impacts on students.

\bibliography{custom}

\appendix

\section{LLM Prompts}\label{sec::llm_prompts}
This section describes the LLM prompt templates used to elicit pairwise preferences from LLMs during the comparison step in Section~\ref{sec::generate_pairwise_comparison}. We design separate LLM prompts for the ASAP and TOEFL11 datasets to reflect their target populations and scoring rubrics. Each LLM prompt includes a system message defining the evaluator's role and a user message with the task context, rubric, and two essays. The model is instructed to return a brief justification and a final decision in structured JSON format for automated parsing. Our LLM prompt format is based on the template introduced by~\citet{lee-etal-2024-unleashing}.
\subsection{ASAP}
\begin{tcolorbox}[title=System Prompt, enhanced, breakable]
   As an English teacher, your primary responsibility is to evaluate the writing quality of essays written by middle school students on an English exam. During the assessment process, you will be provided with a prompt and an essay. First, you should provide comprehensive and concrete feedback that is closely linked to the content of the essay. It is essential to avoid offering generic remarks that could be applied to any piece of writing. \\

   To create a compelling evaluation for both the student and fellow experts, you should reference specific content of the essay to substantiate your assessment. \\

   Next, your task is to determine which essay, Essay 1 or Essay 2, scores higher, or if they score the same, please respond with ``tie''. The evaluation criteria can be part of an overall rubric or separate evaluation criteria. Regardless of the type of rubric, please determine which essay achieves a higher overall score.
\end{tcolorbox}

\begin{tcolorbox}[title=User Prompt, enhanced, breakable]
   \# Prompt

   \{prompt\} \\

   \# Rubric Guidelines

   \{rubric\} \\

   \# Note

   I have made an effort to remove personally identifying information from the essays using the Named Entity Recognizer (NER).

   The relevant entities are identified in the text and then replaced with a string such as ``{PERSON}", ``{ORGANIZATION}", ``{LOCATION}", ``{DATE}", ``{TIME}", ``{MONEY}", ``{PERCENT}”, ``{CAPS}” (any capitalized word) and ``{NUM}” (any digits).
   Please do not penalize the essay because of the anonymizations. \\

   \# Essay1

   \{essay1\} \\

   \# Essay2

   \{essay2\} \\

   Provide your reasoning and final decision in json format:

   \{
      "reasoning": "Your reasoning in one sentence here.",
      "preference": "essay1" or "essay2" or "tie"
   \}
\end{tcolorbox}

\subsection{TOEFL11}
\begin{tcolorbox}[title=System Prompt, enhanced, breakable]
   As an English teacher, your primary responsibility is to evaluate the writing quality of essays written by second language learners on an English exam. During the assessment process, you will be provided with a prompt and an essay. \\

   First, you should provide comprehensive and concrete feedback that is closely linked to the content of the essay. It is essential to avoid offering generic remarks that could be applied to any piece of writing. To create a compelling evaluation for both the student 
   and fellow experts, you should reference specific content of the essay to substantiate your assessment.\\

   Next, your task is to determine which essay, Essay 1 or Essay 2, scores higher, or if they score the same, please respond with ``tie''. The evaluation criteria are based on four assessment categories. Use these categories to comprehensively evaluate and compare the essays, and decide which one achieves a higher overall score.
\end{tcolorbox}

\begin{tcolorbox}[title=User Prompt, enhanced, breakable]
   \# Prompt

   \{prompt\}\\

   \# Rubric Guidelines

   \{rubric\}\\

   \# Essay1

   \{essay1\}\\

   \# Essay2

   \{essay2\}\\

   Provide your reasoning and final decision in json format:

   \{
      "reasoning": "Your reasoning in one sentence here.",
      "preference": "essay1" or "essay2" or "tie"
   \}
\end{tcolorbox}

\section{Embedding Models}\label{sec::embedding_model}

We compare four pretrained embedding models used to convert essays into fixed-length vectors for RankNet. Two of them are OpenAI models: \texttt{text-embedding-3-large} (3072 dimensions) and \texttt{text-embedding-3-small} (1536 dimensions), both of which were designed for semantic similarity tasks. The other two are BERT-base and RoBERTa-base. For these models, we use the [CLS] token from the final hidden layer as the essay representation.

Table~\ref{tab:embedding_models} shows the average QWK scores on ASAP and TOEFL11 using GPT-4o for pairwise comparisons. For ASAP, the choice of embedding model has little impact on performance overall. For TOEFL11, we observe slightly more variation, but all models yield consistently high accuracy. These results suggest that LCES is robust to the choice of embedding encoder.

\begin{table}[t]
   \centering
   \small
   \caption{QWK scores of LCES with different embeddings (using GPT-4o).}
   \label{tab:embedding_models}
   \resizebox{0.48\textwidth}{!}{
   \begin{tabular}{lcc}
   \toprule
   \textbf{Embedding Model} & \textbf{ASAP} & \textbf{TOEFL11} \\
   \midrule
   text-embedding-3-large  & 0.653 & 0.645 \\
   text-embedding-3-small  & 0.668 & 0.630 \\
   BERT-base-uncased       & 0.658 & 0.663 \\
   RoBERTa-base            & 0.655 & 0.601 \\
   \bottomrule
   \end{tabular}
   }
\end{table}

\section{Hyperparameters}\label{sec::hyperparameters}
\begin{table}[t]
   \centering
   \small
   \caption{Hyperparameters for RankNet.}
   \label{tab:ranknet_hyperparameters}
   \begin{tabular}{lc}
   \toprule
   \textbf{Hyperparameter} & \textbf{Value} \\
   \midrule
   Batch size & 4096 \\
   Dropout rate & 0.3 \\
   Hidden units & 256 \\
   Weight decay & 0.01 \\
   \bottomrule
   \end{tabular}
\end{table}
\begin{table*}[t]
   \centering
   \tiny
   \caption{The Spearman rank correlation coefficient for each prompt in ASAP and TOEFL11. \textbf{Bold} indicates the best-performing method for each prompt.}
   \label{tab:asap_toefl_spearman}
   \resizebox{\textwidth}{!}{
   \begin{tabular}{l l l c c c c c c c c c}
       \toprule
       \textbf{Dataset} & \textbf{Model} & \textbf{Method} &
       \textbf{P1} & \textbf{P2} & \textbf{P3} & \textbf{P4} &
       \textbf{P5} & \textbf{P6} & \textbf{P7} & \textbf{P8} & \textbf{Avg.} \\
       \midrule
\multirow{15}{*}{ASAP}
 & Mistral-7B & Vanilla & 0.511 & 0.511 & 0.439 & 0.658 & 0.527 & 0.418 & 0.379 & 0.459 & 0.488 \\
 &                     & MTS     & 0.593 & 0.468 & 0.612 & 0.729 & 0.739 & 0.555 & 0.566 & 0.306 & 0.571 \\
 &                     & LCES    & \textbf{0.616} & \textbf{0.678} & \textbf{0.745} & \textbf{0.784} & \textbf{0.811} & \textbf{0.806} & \textbf{0.684} & \textbf{0.632} & \textbf{0.719} \\
\cmidrule{2-12}
 & Llama-3.2-3B & Vanilla & 0.068 & 0.109 & 0.452 & -0.033 & 0.276 & 0.142 & 0.209 & 0.076 & 0.162 \\
 &                                & MTS     & 0.205 & 0.528 & 0.500 & 0.712 & 0.606 & 0.527 & 0.210 & 0.276 & 0.445 \\
 &                                & LCES    & \textbf{0.665} & \textbf{0.693} & \textbf{0.725} & \textbf{0.767} & \textbf{0.741} & \textbf{0.738} & \textbf{0.589} & \textbf{0.684} & \textbf{0.700} \\
\cmidrule{2-12}
 & Llama-3.1-8B & Vanilla & 0.005 & 0.050 & 0.451 & 0.618 & 0.424 & 0.429 & 0.061 & -0.090 & 0.245 \\
 &                                & MTS     & 0.538 & 0.580 & 0.546 & 0.723 & 0.731 & 0.543 & 0.570 & 0.366 & 0.574 \\
 &                                & LCES    & \textbf{0.702} & \textbf{0.685} & \textbf{0.723} & \textbf{0.809} & \textbf{0.754} & \textbf{0.710} & \textbf{0.724} & \textbf{0.719} & \textbf{0.728} \\
\cmidrule{2-12}
 & GPT-4o-mini & Vanilla & 0.394 & 0.472 & 0.464 & 0.730 & 0.668 & 0.545 & 0.435 & 0.580 & 0.536 \\
 &                               & MTS     & 0.560 & 0.523 & 0.509 & 0.672 & 0.763 & 0.565 & 0.498 & 0.555 & 0.580 \\
 &                               & LCES    & \textbf{0.588} & \textbf{0.678} & \textbf{0.736} & \textbf{0.817} & \textbf{0.761} & \textbf{0.727} & \textbf{0.636} & \textbf{0.693} & \textbf{0.705} \\
\cmidrule{2-12}
 & GPT-4o & Vanilla & 0.468 & 0.518 & 0.525 & 0.787 & 0.729 & 0.557 & 0.546 & 0.549 & 0.585 \\
 &                & MTS     & 0.417 & 0.642 & 0.639 & 0.771 & 0.557 & 0.576 & 0.502 & 0.608 & 0.589 \\
 &                & LCES    & \textbf{0.578} & \textbf{0.682} & \textbf{0.750} & \textbf{0.833} & \textbf{0.812} & \textbf{0.776} & \textbf{0.713} & \textbf{0.713} & \textbf{0.732} \\
       \midrule
\multirow{15}{*}{TOEFL11}
 & Mistral-7B & Vanilla & 0.272 & 0.126 & 0.185 & 0.145 & 0.030 & 0.042 & 0.141 & 0.241 & 0.148 \\
 &                     & MTS     & \textbf{0.717} & \textbf{0.587} & \textbf{0.674} & 0.649 & \textbf{0.703} & \textbf{0.634} & \textbf{0.640} & \textbf{0.740} & \textbf{0.669} \\
 &                     & LCES    & 0.470 & 0.565 & 0.638 & \textbf{0.665} & 0.560 & 0.495 & 0.562 & 0.681 & 0.579 \\
\cmidrule{2-12}
 & Llama-3.2-3B & Vanilla & 0.204 & 0.144 & 0.339 & 0.205 & 0.182 & 0.229 & 0.080 & 0.161 & 0.193 \\
 &                                & MTS     & 0.649 & 0.572 & 0.720 & 0.644 & 0.532 & \textbf{0.549} & 0.608 & 0.563 & 0.604 \\
 &                                & LCES    & \textbf{0.663} & \textbf{0.628} & \textbf{0.748} & \textbf{0.722} & \textbf{0.636} & 0.505 & \textbf{0.627} & \textbf{0.721} & \textbf{0.656} \\
\cmidrule{2-12}
 & Llama-3.1-8B & Vanilla & -0.077 & 0.166 & -0.002 & -0.005 & -0.004 & -0.047 & -0.034 & 0.095 & 0.012 \\
 &                                & MTS     & 0.665 & 0.609 & \textbf{0.791} & 0.686 & 0.647 & 0.542 & 0.622 & 0.663 & 0.653 \\
 &                                & LCES    & \textbf{0.751} & \textbf{0.668} & 0.759 & \textbf{0.755} & \textbf{0.723} & \textbf{0.582} & \textbf{0.690} & \textbf{0.767} & \textbf{0.712} \\
\cmidrule{2-12}
 & GPT-4o-mini & Vanilla & 0.131 & 0.252 & 0.261 & 0.172 & 0.044 & 0.123 & 0.151 & 0.177 & 0.164 \\
 &                               & MTS     & 0.684 & 0.655 & \textbf{0.781} & 0.716 & 0.727 & 0.645 & 0.650 & 0.715 & 0.696 \\
 &                               & LCES    & \textbf{0.753} & \textbf{0.674} & 0.757 & \textbf{0.745} & \textbf{0.753} & \textbf{0.684} & \textbf{0.695} & \textbf{0.769} & \textbf{0.729} \\
\cmidrule{2-12}
 & GPT-4o & Vanilla & 0.257 & 0.244 & 0.440 & 0.239 & 0.253 & 0.270 & 0.258 & 0.323 & 0.285 \\
 &                & MTS     & 0.675 & 0.655 & \textbf{0.802} & 0.713 & 0.728 & \textbf{0.628} & 0.635 & 0.727 & 0.695 \\
 &                & LCES    & \textbf{0.748} & \textbf{0.712} & 0.768 & \textbf{0.733} & \textbf{0.779} & 0.614 & \textbf{0.699} & \textbf{0.784} & \textbf{0.730} \\
       \bottomrule
   \end{tabular}}
\end{table*}
Table~\ref{tab:ranknet_hyperparameters} shows the hyperparameters used for training the RankNet model described in Section~\ref{sec::ranknet}. The model consists of two linear layers with a ReLU activation and a dropout layer applied between them. Weight decay is applied as part of the Adam optimizer configuration.

\section{Evaluation by Spearman Rank Correlation Coefficient}\label{sec::spearman_correlation}
In addition to the primary metric QWK, we report Spearman rank correlation coefficients to evaluate the ordinal consistency between predicted and gold-standard scores. This metric is especially relevant in applications where preserving the relative ranking of essays is more important than matching exact scores. Compared with the baseline methods, LCES generally achieves higher Spearman correlations across most prompts and LLMs (Table~\ref{tab:asap_toefl_spearman}), supporting its strength in maintaining rank order.
\begin{table}[t]
   \centering
   \small
   \caption{Agreement rates (\%) between LLMs and human evaluators in pairwise comparisons.}
   \label{tab:agreement_rates}
   \begin{tabular}{lcccc}
   \toprule
   \multirow{2}{*}{\textbf{Model}} & \multicolumn{2}{c}{\textbf{ASAP}} & \multicolumn{2}{c}{\textbf{TOEFL11}} \\
   \cmidrule(lr){2-3} \cmidrule(lr){4-5}
   & \textbf{All} & \textbf{Excl. Ties} & \textbf{All} & \textbf{Excl. Ties} \\
   \midrule
   Mistral-7B       & 55.9 & 58.0 & 52.1 & 41.2 \\
   Llama-3.2-3B     & 56.3 & 65.0 & 54.5 & 60.1 \\
   Llama-3.1-8B     & 60.3 & 71.6 & 57.6 & 76.6 \\
   GPT-4o-mini      & 59.9 & 75.1 & 55.9 & 86.6 \\
   GPT-4o           & 64.3 & 80.0 & 57.8 & 83.0 \\
   \bottomrule
   \end{tabular}
\end{table}

\section{Agreement Rate}\label{sec::agreement_rate}

To further validate the reliability of LLM-generated pairwise comparisons, we measure the agreement rate between LLM decisions and human annotations on a subset of evaluation pairs. We report results for two metrics (Table~\ref{tab:agreement_rates}): \textbf{All}, which reflects agreement across all pairs including ties, and \textbf{Excl. Ties}, which excludes cases where the gold-standard label indicates a tie. The latter focuses on pairs where a clear score difference exists and thus better captures the LLM's ability to detect meaningful distinctions.

Better-performing LLMs such as GPT-4o and Llama-3.1-8B show higher agreement rates, particularly when ties are excluded. These results are consistent with the final scoring performance in terms of both QWK and Spearman correlation, supporting the use of agreement rate as an indicator of pairwise comparison quality.

\end{document}